\documentclass[conference]{IEEEtran}
\IEEEoverridecommandlockouts

\usepackage{cite}
\usepackage{amsmath,amssymb,amsfonts}
\usepackage{algorithmic}
\usepackage{graphicx}
\usepackage{textcomp}
\usepackage{xcolor}

\usepackage{subfigure}
\usepackage{multirow}
\usepackage[font=small]{caption}
\usepackage[ruled,vlined]{algorithm2e}
\usepackage{bm}

\def\BibTeX{{\rm B\kern-.05em{\sc i\kern-.025em b}\kern-.08em
    T\kern-.1667em\lower.7ex\hbox{E}\kern-.125emX}}
\begin{document}

\title{InfraredGP: Efficient Graph Partitioning via Spectral Graph Neural Networks with Negative Corrections
\thanks{This paper serves as an advanced extension of our prior work \cite{qin2025efficient}. It introduces a modified method with novel designs and reports new results for both static and streaming GP on the IEEE HPEC Graph Challenge benchmark.}
}

\author{\IEEEauthorblockN{Meng Qin\IEEEauthorrefmark{2}, Weihua Li\IEEEauthorrefmark{2}\IEEEauthorrefmark{3}\IEEEauthorrefmark{4}\IEEEauthorrefmark{1}, Jinqiang Cui\IEEEauthorrefmark{2}\IEEEauthorrefmark{1}, Sen Pei\IEEEauthorrefmark{5}}
\IEEEauthorblockA{
\IEEEauthorrefmark{2}Department of Strategic and Advanced Interdisciplinary Research, Pengcheng Laboratory (PCL)\\
\IEEEauthorrefmark{3}LMIB, NLSDE, BDBC, and School of Artificial Intelligence, Beihang University\\
\IEEEauthorrefmark{4}Hangzhou International Innovation Institute, Beihang University\\
\IEEEauthorrefmark{5}Department of Environmental Health Sciences, Mailman School of Public Health, Columbia University\\
\IEEEauthorrefmark{1}Corresponding Authors: Weihua Li (weihuali@buaa.edu.cn), Jinqiang Cui (cuijq@pcl.ac.cn)
\\(\textbf{Honorable Mention} of IEEE HPEC 2025 Graph Challenge: https://graphchallenge.mit.edu/champions)
}
}

\maketitle

\begin{abstract}
Graph partitioning (GP), a.k.a. community detection, is a classic problem that divides nodes of a graph into densely-connected blocks. From a perspective of graph signal processing, we find that graph Laplacian with a negative correction can derive graph frequencies beyond the conventional range $[0, 2]$. To explore whether the low-frequency information beyond this range can encode more informative properties about community structures, we propose InfraredGP. It (\romannumeral1) adopts a spectral GNN as its backbone combined with low-pass filters and a negative correction mechanism, (\romannumeral2) only feeds random inputs to this backbone, (\romannumeral3) derives graph embeddings via one feed-forward propagation (FFP) without any training, and (\romannumeral4) obtains feasible GP results by feeding the derived embeddings to BIRCH. Surprisingly, our experiments demonstrate that based solely on the negative correction mechanism that amplifies low-frequency information beyond $[0, 2]$, InfraredGP can derive distinguishable embeddings for some standard clustering modules (e.g., BIRCH) and obtain high-quality results for GP without any training. Following the IEEE HPEC Graph Challenge benchmark, we evaluate InfraredGP for both static and streaming GP, where InfraredGP can achieve much better efficiency (e.g., 16x-23x faster) and competitive quality over various baselines. We have made our code public at https://github.com/KuroginQin/InfraredGP
\end{abstract}


\section{Introduction}\label{Sec:Intro}
Graph partitioning (GP), a.k.a. community detection \cite{fortunato202220}, is a classic problem that partitions nodes of a graph into densely connected blocks (i.e., communities or clusters). As the extracted blocks may correspond to vital substructures of real-world systems (e.g., functional groups in protein-protein interactions \cite{Berahmand}), many network applications are thus formulated as GP (e.g., cellular network decomposition \cite{dai2017optimal}, parallel task assignment \cite{hendrickson2000graph}, and Internet traffic profiling \cite{qin2019towards}).

Focusing on the NP-hard nature of GP, which implies the difficulty in balancing efficiency and quality of this problem, the IEEE HPEC Graph Challenge \cite{kao2017streaming} provides a competitive benchmark to evaluate both aspects. It has so far attracted various solutions, including fast approximation of the stochastic block model (SBM) with sampling \cite{wanye2019fast}, batching \cite{wanye2023integrated}, and Kalman filter \cite{durbeck2022kalman}; incremental spectral clustering with LOBPCG \cite{zhuzhunashvili2017preconditioned}; community-preserving graph embedding with randomized \cite{gao2023raftgp} and pre-trained \cite{qin2024towards} graph neural networks (GNNs). Different from these solutions, this study presents an alternative method based on the following investigation from a perspective of graph signal processing (GSP).

\begin{figure*}
\centering
 \begin{minipage}{0.325\linewidth}
 \subfigure[Eigenvector ${\bf{u}}_0$, $\tau = 0$]{
  \includegraphics[width=\textwidth,trim=40 0 19 22,clip]{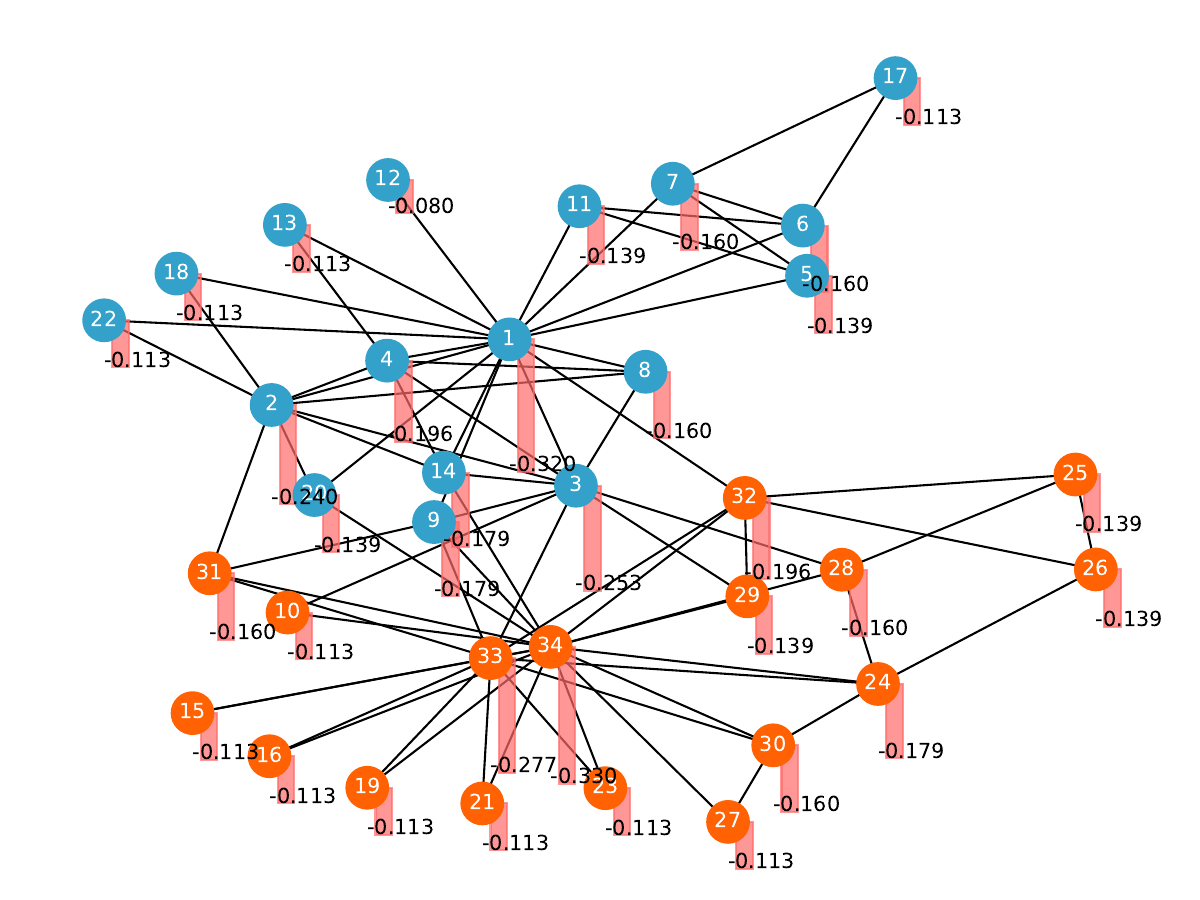}}
 \end{minipage}
 \begin{minipage}{0.325\linewidth}
 \subfigure[${\bf{u}}_1$ (i.e., low-frequency info.), $\tau = 0$]{
  \includegraphics[width=\textwidth,trim=40 0 20 22,clip]{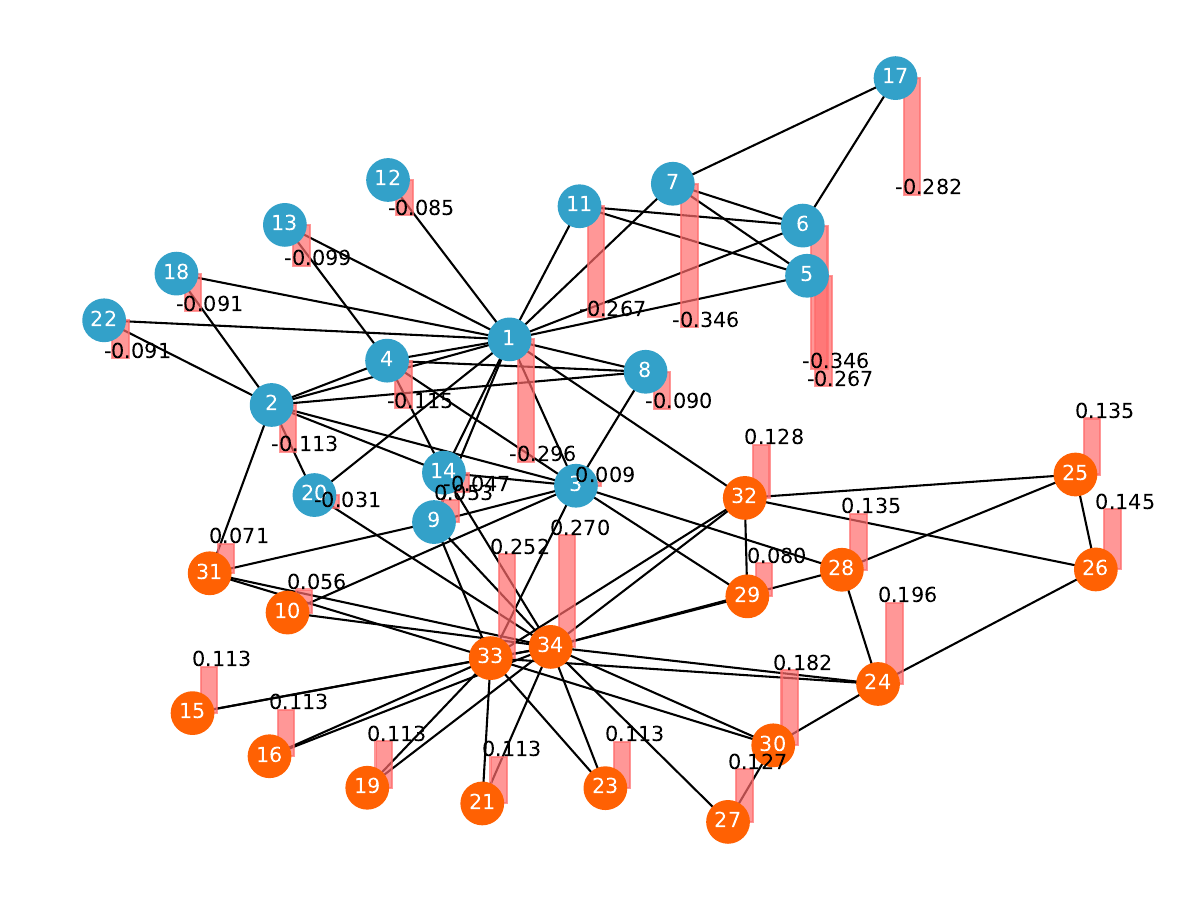}}
 \end{minipage}
 \begin{minipage}{0.325\linewidth}
 \subfigure[${\bf{u}}_{33}$ (i.e., high-frequency info.), $\tau = 0$]{
  \includegraphics[width=\textwidth,trim=40 0 20 22,clip]{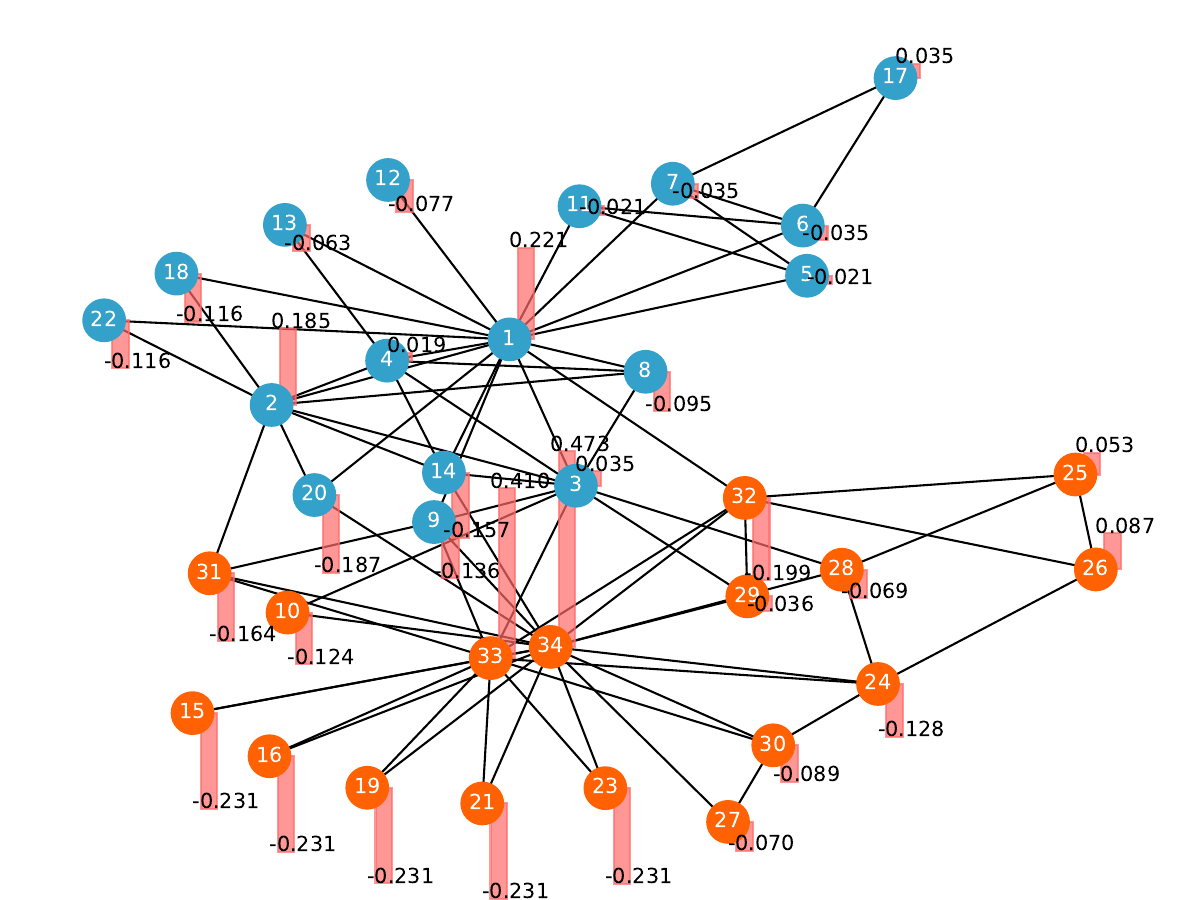}}
 \end{minipage}
 \begin{minipage}{0.17\linewidth}
 \subfigure[$\{ \tilde \lambda_r \}$, $\tau=-1.5$]{
  \includegraphics[width=\textwidth,trim=5 28 25 18,clip]{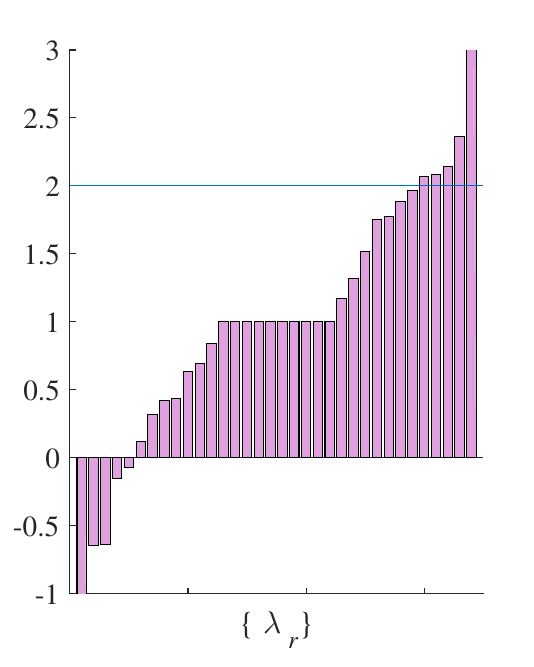}}
 \end{minipage}
 \begin{minipage}{0.17\linewidth}
 \subfigure[$\{ \tilde \lambda_r \}$, $\tau=-1.0$]{
  \includegraphics[width=\textwidth,trim=5 28 25 18,clip]{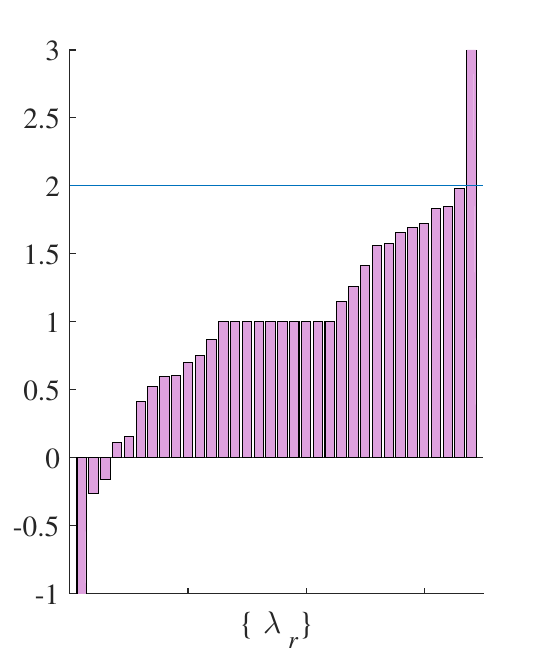}}
 \end{minipage}
 \begin{minipage}{0.235\linewidth}
 \subfigure[$\{ \lambda_r \}$, $\tau=0$]{
  \includegraphics[width=\textwidth,trim=23 40 22 30,clip]{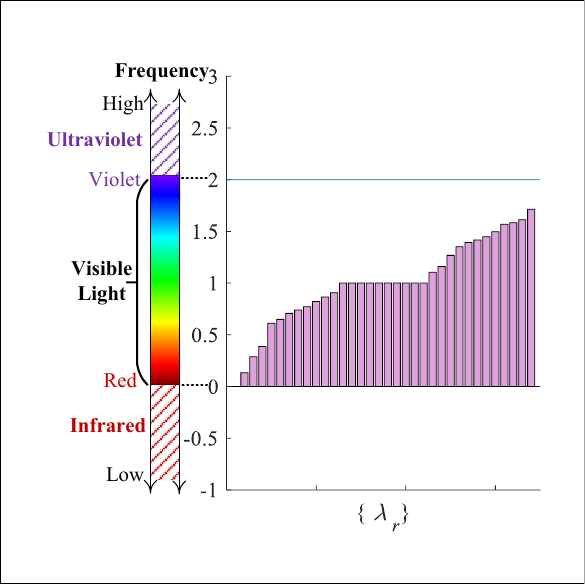}}
 \end{minipage}
 \begin{minipage}{0.17\linewidth}
 \subfigure[$\{ \tilde \lambda_r \}$, $\tau=5$]{
  \includegraphics[width=\textwidth,trim=5 28 25 18,clip]{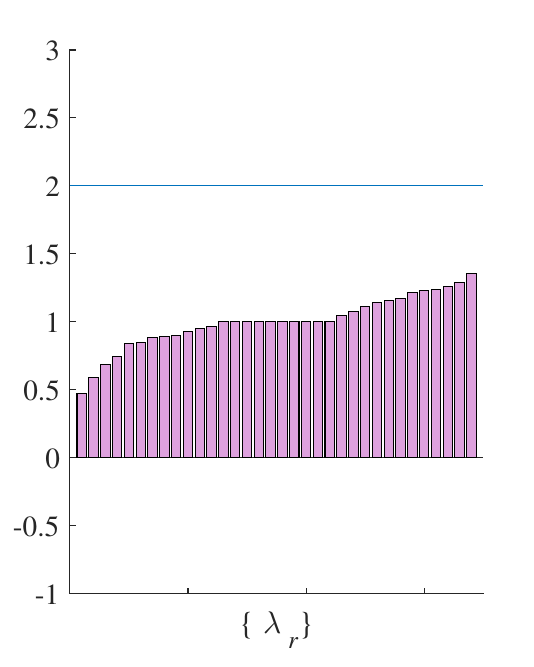}}
 \end{minipage}
 \begin{minipage}{0.17\linewidth}
 \subfigure[$\{ \tilde \lambda_r \}$, $\tau=10$]{
  \includegraphics[width=\textwidth,trim=5 28 25 18,clip]{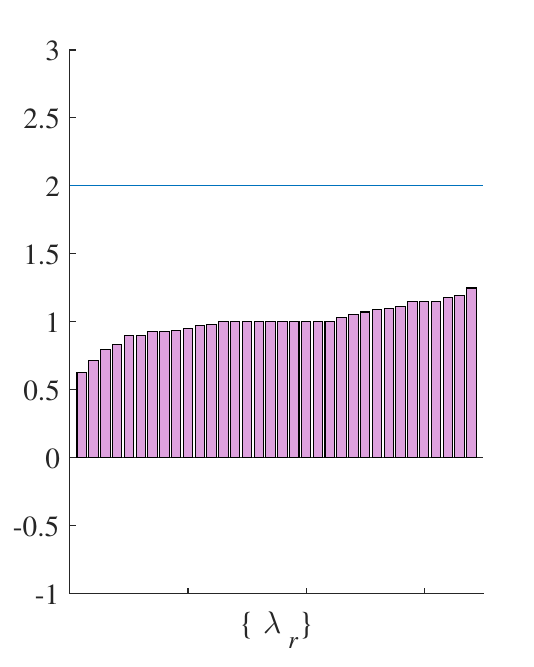}}
 \end{minipage}
 \vspace{-0.2cm}
\caption{An example about (\textbf{a}-\textbf{c},\textbf{f}) graph spectrum (i.e., ED on graph Laplacian ${\bf{L}}$ with $\tau = 0$) and (\textbf{d},\textbf{e},\textbf{g},\textbf{h}) effects of degree correction (i.e., $\tau >0$ and $< 0$) on the \textbf{Karate Club} dataset with $34$ nodes. Each color denotes a unique block member.
}\label{Fig:Toy}
\vspace{-0.4cm}
\end{figure*}

\textbf{Investigation from a GSP View}.
Let ${\bf{A}}$ be the adjacency matrix of a graph with $N$ nodes. In GSP, we first get the graph Laplacian ${\bf{L}} := {\bf{I}} - {\bf{D}}^{-1/2}{\bf{A}}{\bf{D}}^{-1/2}$ and conduct eigen-decomposition (ED) on ${\bf{L}}$, with ${\bf{I}}$ and ${\bf{D}}$ as an identity matrix and degree diagonal matrix w.r.t. ${\bf{A}}$. The eigenvalues $\{ \lambda_s \}$ of ${\bf{L}}$, which can be arranged in the order $0 = \lambda_0 \le \lambda_1 \le \cdots \le \lambda_{N-1} \le 2$, are defined as \textit{\textbf{frequencies}} of the graph. Let ${\bf{u}}_s$ be the eigenvector of $\lambda_s$. The \textit{\textbf{graph frequency}} $\lambda_s$ describes the variation of corresponding values in ${\bf{u}}_s$ w.r.t. local topology.

Fig.~\ref{Fig:Toy} (f) and (a-c) visualize (\romannumeral1) the distribution of eigenvalues $\{ \lambda_s \}$ and (\romannumeral2) major eigenvectors $\{ {\bf{u}}_0, {\bf{u}}_1, {\bf{u}}_{33} \}$ for the well-known \textbf{Karate Club} dataset \cite{zachary1977information} with $34$ nodes and $2$ blocks, where each color denotes a unique block member.
Concretely, all the values in ${\bf{u}}_0$ w.r.t. $\lambda_0$ are negative, indicating that there is no significant variation w.r.t. local topology (i.e., \textit{\textbf{zero frequency}}).
For $\lambda_1$ (i.e., \textit{\textbf{the smallest non-zero frequency}}\footnote{The multiplicity of zero eigenvalue equals to the number of connected components \cite{shuman2013emerging}. A connected graph only has one connected component.}), one can extract the block membership based on signs of entries in ${\bf{u}}_1$, where nodes in the blue (orange) block are more likely to have negative (positive) values. Some nodes (e.g., $\{ 3, 9, 14, 20\}$) are with values close to $0$, as it is hard to determine which block they belong to. In contrast, entries in ${\bf{u}}_{33}$ w.r.t. $\lambda_{33}$ (i.e., \textit{\textbf{the highest frequency}}) have significant variation w.r.t. local topology. For instance, there is a sign change between nodes $34$ and $23$ (i.e., $({\bf{u}}_{33})_{34} = 0.473 > 0$ but $({\bf{u}}_{33})_{23} = -0.231 < 0$), even though they are directly connected and in the same block. 

In summary, \textit{low-frequency information in the spectral domain is more likely to encode intrinsic community structures for GP}, consistent with motivations of the well-known spectral clustering \cite{von2007tutorial}.
Our prior work \cite{qin2025efficient} also validated that \textit{high-frequency information may encode structural roles \cite{yan2024pacer,mengirwe} of nodes in the topology} (see \cite{qin2025efficient} for details of this property).

Inspired by the degree corrected spectral clustering (DCSC), our prior study \cite{qin2025efficient} further demonstrated that ED on graph Laplacian ${\bf{L}}_{\tau} := {\bf{I}} - {\bf{D}}_{\tau}^{-1/2}{\bf{A}}{\bf{D}}_{\tau}^{-1/2}$ (with ${\bf{D}}_{\tau} := {\bf{D}} + \tau {\bf{I}}$), which involves a degree correction term $\tau > 0$, may help derive more informative graph embeddings (i.e., low-dimensional vector representations for nodes). Let $\{ \tilde \lambda_s \}$ and $\{ {\bf{\tilde u}}_s \}$ be eigenvalues and eigenvectors of ${\bf{L}}_{\tau}$.
The following Gershgorin Circle Theorem \cite{barany2017gershgorin} helps interpret the effect of $\tau$.

\textbf{Theorem~1} (\textbf{Gershgorin Circle Theorem}). Eigenvalues of ${\bf{M}} \in \mathbb{C}^{N \times N}$ lie in the union of $N$ discs $(\mathcal{D}_1, \cdots, \mathcal{D}_N)$, where $\mathcal{D}_i := x \in \mathbb{C} : |x - {\bf{M}}_{ii}| \le r_i$; $r_i := \sum\nolimits_{j = 1,j \ne i}^N {|{{\bf{M}}_{ij}}|}$.

It indicates that eigenvalues of ${\bf{I}}_N - {\bf{D}}_{\tau}^{-1} {\bf{A}}$ (analogous to ${\bf{I}}_N - {\bf{D}}_{\tau}^{-1/2} {\bf{A}} {\bf{D}}_{\tau}^{-1/2}$) lie in circles centered at $1$, whose radiuses are controlled by $\{ {\deg}_i / ({\deg}_i + \tau) \}$, with ${\deg}_i$ as the degree of node $v_i$.
Namely, $\tau$ adjusts the distribution of eigenvalues (i.e., graph frequencies) $\{ \tilde \lambda_r \}$.
As illustrated in Fig.~\ref{Fig:Toy} (f-h), the increase of $\tau$ forces graph frequencies $\{ \tilde \lambda_s \}$ to approach $1$, verifying the aforementioned interpretation that $\{ \tilde \lambda_s \}$ lie in circles centered at $1$ with radiuses ${\mathop {\lim }\limits_{\tau  \to  + \infty }}{\deg _i}/({\deg _i} + \tau ) = 0$.

Related studies usually consider the standard graph Laplacian ${\bf{L}}$ (i.e., $\tau = 0$) or ${\bf{L}}_\tau$ with $\tau > 0$, where eigenvalues must be within $[0, 2]$, analogous to the spectrum of visible light in optics as highlighted in Fig.~\ref{Fig:Toy} (f).
Here, we consider another extreme with $\tau < 0$. To avoid infinite or negative radiuses of circles in \textbf{Theorem~1}, we let
\begin{equation}\label{Eq:Neg-tau}
    {({{\bf{D}}_\tau })_{ii}} = {\deg _i} - \min \{ |\tau |,{\deg _i} - \epsilon \},
\end{equation}
when deriving ${\bf{D}}_\tau$ and ${\bf{L}}_\tau$ for $\tau < 0$, with $\epsilon$ as a small positive value (e.g., $\epsilon = 0.001$ in our experiment).

Surprisingly, as shown in Fig.~\ref{Fig:Toy} (d-e), the negative correction (i.e., $\tau  < 0$) can derive graph frequencies $\{\tilde \lambda_s\}$ out of the conventional range $[0, 2]$. Analogous to the optical spectrum, we define $\{ ({{\tilde \lambda }_s},{{{\bf{\tilde u}}}_s})|{{\tilde \lambda }_s} < 0\}$ (or $\{ ({{\tilde \lambda }_s},{{{\bf{\tilde u}}}_s})|{{\tilde \lambda }_s} > 2\}$) as the  \textit{\textbf{infrared}} (or \textit{\textbf{ultraviolet}}) \textit{\textbf{information}} in graph spectrum.

\textbf{Presented Work}.
In this study, we investigate whether the infrared information derived by $\tau < 0$ can encode more informative properties about community structures, compared with those of $\tau = 0$ and $\tau > 0$.
As an advanced extension of our prior work \cite{qin2025efficient}, we propose InfraredGP, a simple yet effective method for the $K$-agnostic GP problem in Graph Challenge, where the number of blocks (i.e., clusters) $K$ is not given and should be determined by a GP algorithm.

InfraredGP adopts a spectral GNN as its backbone, only uses random noises as inputs, and derives low-dimensional embeddings via just one feed-forward propagation (FFP). The derived embeddings are then fed into BIRCH \cite{zhang1996birch}, an efficient $K$-agnostic clustering algorithm, to derive feasible GP results.

In particular, we apply a negative correction mechanism (i.e., with $\tau < 0$) to each GNN layer and combine it with a simple low-pass filter that amplifies low-frequency information, especially the infrared information.
Surprisingly, our experiments demonstrate that only based on this architecture with $\tau < 0$, our method can derive distinguishable embeddings for $K$-agnostic GP and obtain high-quality results. As there are no additional procedures of feature extraction, model training, and ED, InfraredGP is also very efficient.

The major contributions of this paper beyond our prior study \cite{qin2025efficient} are summarized as follows.

\begin{itemize}
    \item To the best of our knowledge, we are the first to explore the infrared information via a negative correction mechanism. Whereas, our prior method only considers graph frequencies $\{ \tilde \lambda_s \}$ within $[0, 2]$. This paper is also the first GSP-based solution submitted to Graph Challenge.
    \item We evaluate the ability of InfraredGP to support both static and streaming $K$-agnostic GP on the Graph Challenge benchmark, where our method can achieve much better efficiency and competitive quality over various baselines. In contrast, our prior method cannot directly support $K$-agnostic GP.
\end{itemize}

\section{Problem Statement \& Preliminaries}\label{Sec:Prob}
We follow the IEEE HPEC Graph Challenge \cite{kao2017streaming} to consider GP on undirected, unweighted, and connected graphs. A graph can be represented as $G := (V, E)$, with $V := \{ v_1, v_2, \cdots, v_N\}$ and $E := \{ (v_i, v_j) | v_i, v_j \in V\}$ as the sets of nodes and edges.
The topology structure of $G$ can be described by an adjacency matrix ${\bf{A}} \in \{ 0, 1\}^{N \times N}$, where  ${\bf{A}}_{ij} = {\bf{A}}_{ji} = 1$ if $(v_i, v_j) \in E$ and ${\bf{A}}_{ij} = {\bf{A}}_{ji} = 0$, otherwise.

(\textbf{Static GP}). The static GP aims to partition $V$ into $K$ disjoint subsets $C := (C_1, \cdots, C_K)$ (i.e., blocks, communities, or clusters) s.t. (\romannumeral1) linkages within each block are dense but (\romannumeral2) those between blocks are relatively loose. In particular, Graph Challenge considers the $K$-agnostic GP, where the number of blocks $K$ is unknown. One should simultaneously determine $K$ and the corresponding block partition $C$.

(\textbf{Streaming GP}). The streaming GP serves as an advanced extension of the static setting. Graph Challenge provides two alternative models \cite{kao2017streaming} for this extension. We consider the more challenging snowball model and leave the merging edge model for future work. The snowball model first divides $V$ into $T$ disjoint subsets $(V_1, \cdots, V_T)$, with $V_t$ as the set of newly added nodes in the $t$-th step. Let $\tilde V_t := \bigcup\nolimits_{t = 1}^r {{V_t}} $ and $\tilde E_t$ be the set of edges induced by $\tilde V_r$. For each step $t$, streaming GP derives a $K$-agnostic block partition based on $(\tilde V_t, \tilde E_t)$.

For eigenvalues (i.e., graph frequencies) in the order $\lambda_0 \le \lambda_1 \le \cdots \le \lambda_{N-1}$, we arrange them and their eigenvectors in matrices ${\bf{\Lambda}} :={\mathop{\rm diag}\nolimits} ({\lambda _0},{\lambda _1}, \cdots, \lambda_{N-1} )$ and ${\bf{U}} := [{\bf{u}}_0, {\bf{u}}_1, \cdots, {\bf{u}}_{N-1}]$. Then, ED on ${\bf{L}}$ and ${\bf{L}}_{\tau}$ can be represented as ${\bf{L}} = {\bf{U}} {\bf{\Lambda}} {\bf{U}}^T$ and ${\bf{L}}_{\tau} = {\bf{\tilde U}} {\bf{\tilde \Lambda}} {\bf{\tilde U}}^T$.

(\textbf{Graph Signal Processing}). Given a signal ${\bf{z}} \in \mathbb{R}^{N}$ to be processed on graph $G$, the \textit{\textbf{graph Fourier transform}} is defined as ${\bf{\hat z}} = \mathcal{F} ({\bf{z}}) := {\bf{U}}^T {\bf{z}}$. It uses orthogonal $\{ {\bf{u}}_i \}$ as bases and maps ${\bf{z}}$ from the spatial to spectral domains. The \textit{\textbf{inverse Fourier transform}} is defined as ${\bf{z}} = \mathcal{F}^{-1} ({\bf{\hat z}}) := {\bf{U}} {\bf{\hat z}}$ that recovers ${\bf{\hat z}}$ from the spectral to spatial domains. Let $\varphi ({\bf{\Lambda}})$ be the \textit{\textbf{graph convolutional kernel}}, a pre-defined or learnable function about ${\bf{\Lambda}}$. The \textit{\textbf{graph convolution}} operation can then be represented as $\varphi * {\bf{z}} := {\bf{U}} \varphi ({\bf{\Lambda}}) {\bf{U}}^{T} {\bf{z}}$.

\section{Methodology}\label{Sec:Meth}
Based on our investigation of the infrared information with $\tau < 0$ in Section~\ref{Sec:Intro}, we propose InfraredGP for both static and streaming GP. Concretely, it (\romannumeral1) adopts a spectral GNN as its backbone combined with low-pass filters and a negative correction mechanism, (\romannumeral2) only uses random noises as inputs, and (\romannumeral3) derives graph embeddings via just one FFP.
InfraredGP is thus efficient without relying on any procedures of feature extraction, model training, and ED.
We then obtain feasible GP results by feeding the derived embeddings to BIRCH \cite{zhang1996birch}, an efficient $K$-agnostic clustering algorithm.

\subsection{Model Architecture}
Motivated by the spectral GNNs in \cite{bo2021beyond,dong2021adagnn}, we consider the following kernel (i.e., filter) w.r.t. the ED on ${\bf{L}}_{\tau}$
\begin{equation}\label{Eq:Filter}
    \varphi ({\bf{\tilde \Lambda }}): = (\theta  + \alpha ){{\bf{I}}_N} - \alpha {\bf{\tilde \Lambda }},
\end{equation}
where $\alpha \in [-1, 1]$ and $\theta \in [0, 1]$ are tunable parameters.
In particular, the sign of $\alpha$ determines $\varphi ({\bf{\tilde \Lambda}})$ is a \textit{\textbf{low-}} or \textit{\textbf{high-pass filter}}.
When $\alpha > 0$, $\varphi ({\bf{\tilde \Lambda}})$ is a monotone decreasing function about ${\bf{\tilde \Lambda}}$, which gives larger weights for bases $\{ \bf{\tilde u}_s \}$ w.r.t. low graph frequencies $\{ \tilde \lambda_s \}$ (i.e., amplifying low-frequency information), and thus represents a low-pass filter.
Similarly, when $\alpha < 0$, $\varphi ({\bf{\tilde \Lambda}})$ becomes a high-pass filter amplifying high-frequency information.

As the low-frequency information encodes more information about community structures, InfraredGP adopts a simple low-pass filter with $(\alpha, \theta) = (1, 0.1)$.
By using the orthogonal property that ${\bf{\tilde U}}{\bf{\tilde U}}^T = {\bf{\tilde U}}^T{\bf{\tilde U}} = {\bf{I}}$, the graph convolutional operation can be formulated as
\begin{equation}\label{Eq:Conv}
    \varphi  * {\bf{Z}} = {\bf{\tilde U}}\varphi ({\bf{\tilde \Lambda }}){{\bf{\tilde U}}^T}{\bf{Z}} = 0.1 {\bf{Z}} + {{\bf{D}}_{\tau}^{ - 1/2}}{\bf{A}}{{\bf{D}}_{\tau}^{ - 1/2}}{\bf{Z}},
\end{equation}
where the application of low-pass filter is equivalent to deriving the sum between original signal $0.1{\bf{Z}}$ and the `mean' value ${{\bf{D}}_{\tau}^{ - 1/2}}{\bf{A}}{{\bf{D}}_{\tau}^{ - 1/2}}{\bf{Z}}$ w.r.t. $1$-hop neighbors, without conducting ED on ${\bf{L}}_{\tau}$.
Different from our prior work \cite{qin2025efficient} and DCSC with $\tau > 0$, InfraredGP considers $\tau < 0$ and applies (\ref{Eq:Neg-tau}) to (\ref{Eq:Conv}), which amplifies the infrared information.
We treat (\ref{Eq:Conv}) as the convolutional operation of one GNN layer and construct the backbone of InfraredGP by stacking multiple such layers.

Let ${\bf{Z}}^{(l-1)}$ and ${\bf{Z}}^{(l)}$ be the input and output of the $l$-th layer. As we only feed random noise to the backbone, we set ${\bf{Z}}^{(0)} \in \mathbb{R}^{N \times d} \sim \mathcal{N} (0, 1/d)$, with $d$ as the embedding dimensionality.
The $l$-th layer is defined as
\begin{equation}\label{Eq:Layer}
    {\bf{Z}}^{(l)} = {\mathop{\rm ZNorm}\nolimits} (\tanh (\varphi  * {{\bf{Z}}^{(l - 1)}})) {\rm{~and~}} {\bf{\tilde Z}} = \sigma ({{\bf{Z}}^{(L)}}),
\end{equation}
where ${\mathop{\rm ZNorm}\nolimits} ({\bf{x}}) := ({\bf{x}} - \mu ({\bf{x}}))/s({\bf{x}})$ denotes the (column-wise) z-score normalization, with $\mu ({\bf{x}})$ and $s({\bf{x}})$ as the mean and standard derivation of entries in ${\bf{x}}$; $\sigma (x): = {(1 + {e^{ - x}})^{ - 1}}$ is the sigmoid function. Suppose that there are $L$ layers. In (\ref{Eq:Layer}), we use ${\bf{\tilde Z}} \in \mathbb{R}^{N \times d}$ to represent the final embeddings derived by InfraredGP, with ${\bf{\tilde Z}}_{i, :}$ as the embedding of node $v_i$.

Note that the normalization ${\mathop{\rm ZNorm}\nolimits} (\cdot)$ and nonlinear activation (i.e., $\tanh (\cdot)$ and $\sigma (\cdot)$) are necessary for InfraredGP to derive informative embeddings. Otherwise, InfraredGP may suffer from significant quality degradation.
A possible interpretation is that these operations can control the magnitude of embeddings to avoid numerical overflow.
For instance, outputs of $\tanh (\cdot)$ and $\sigma (\cdot)$ are within $[-1, 1]$ and $[0, 1]$, which prevents embeddings from being extremely large. ${\mathop{\rm ZNorm}\nolimits} (\cdot)$ is also a commonly used technique to alleviate the impact brought by the magnitude difference of data.

Given the derived embeddings ${\bf{\tilde Z}}$, we obtain a feasible $K$-agnostic GP result by feeding ${\bf{\tilde Z}}$ to BIRCH \cite{zhang1996birch}, an efficient clustering algorithm without relying on a pre-set number of clusters (i.e., blocks) $K$.

Surprisingly, our experiments demonstrate that only based on the negative correction mechanism (i.e., with $\tau < 0$) that amplifies the infrared information, our method can derive distinguishable embeddings for some standard clustering modules (e.g., BIRCH) via one FFP and obtain high-quality results for $K$-agnostic GP.
In contrast, other settings w.r.t. $\tau = 0$ or $\tau >0$ suffer from significant quality degradation.

\subsection{Standard Algorithm for Static GP}

\begin{algorithm}[t]\small
\caption{\small Standard Algorithm for Static GP}
\label{Alg:Static}
\LinesNumbered
\KwIn{static topology $(V, E)$; negative correction $\tau < 0$; number of layers $L$; embedding dimensionality $d$}
\KwOut{$K$-agnostic partition $C$ w.r.t. $(V, E)$}
get ${\bf{D}}_{\tau}$ with negative correction $\tau  < 0$ via (\ref{Eq:Neg-tau})\\
generate random noise via $\bm{\Theta} \in \mathbb{R}^{N \times d} \sim \mathcal{N} (0, 1/d)$\\
${{\bf{Z}}^{(0)}} \leftarrow {\bm{\Theta}} $ \\
\For{$l$ \bf{from} $1$ \bf{to} $L$}
{
    ${\bf{Z}}^{(l)} \leftarrow {\mathop{\rm ZNorm}\nolimits} (\tanh (\varphi  * {{\bf{Z}}^{(l - 1)}}))$ using (\ref{Eq:Conv}) and (\ref{Eq:Layer})\\
}
${\bf{\tilde Z}} \leftarrow \sigma ({{\bf{Z}}^{(L)}})$ using (\ref{Eq:Layer}) \\
fit BIRCH to ${\bf{\tilde Z}}$ and derive $K$-agnostic clustering result $C$
\end{algorithm}

Algorithm~\ref{Alg:Static} summarizes the standard procedure of static GP, where $L$ and $d$ are the number of GNN layers and embedding dimensionality; $\tau < 0$ is a pre-set negative correction term.
Concretely, we first pre-compute the degree diagonal matrix ${\bf{D}}_{\tau}$ (i.e., line 1) and generate Gaussian random noise ${\bm{\Theta}}$ (i.e., line 2). We then feed ${\bm{\Theta}}$ to the GNN backbone and derive embeddings ${\bf{\tilde Z}}$ via one FFP (i.e., lines 3-6), where each GNN layer serves as one iteration to update embeddings. One can obtain a feasible $K$-agnostic GP result $C$ by fitting the BIRCH clustering algorithm to ${\bf{\tilde Z}}$ (i.e., line 7).

Let $N$ and $M$ be the numbers of nodes and edges. For a large sparse graph, we assume that $L < d \ll N < M$. The complexity of pre-computing degree diagonal matrix ${\bf{D}}_{\tau}$ and generating random input ${\bm{\Theta}}$ is $O (N + Nd) \approx O(N)$.
Based on the efficient sparse-dense matrix multiplication, the complexity of one iteration is $O ((N +M)d) \approx O(N +M)$.
For a backbone with $L$ layers (i.e., iterations), the complexity of one FFP is $O((N + M)L) \approx O(N +M)$.
Moreover, the complexity of BIRCH is approximately $O(N)$ \cite{zhang1996birch}. 
In summary, the overall complexity of Algorithm~\ref{Alg:Static} is no more than $O (N + M)$.

\subsection{Extension to Streaming GP}

\begin{algorithm}[t]\small
\caption{\small Extended Algorithm for Streaming GP}
\label{Alg:Stream}
\LinesNumbered
\KwIn{streaming topology $(V_t, E_t)$ of the $t$-th step; previous cumulative topology $({\tilde V}_{t-1}, {\tilde E}_{t-1})$; embeddings ${\bf{\hat Z}}_{t-1}$ in previous step; negative correction $\tau < 0$; number of layers $L$; embedding dimensionality $d$}
\KwOut{$K$-agnostic partition $C^{(t)}$ w.r.t. cumulative $({\tilde V}_t, {\tilde E}_t)$}
get $({\tilde V}_t, {\tilde E}_t)$ by merging $({\tilde V}_{t-1}, {\tilde E}_{t-1})$ and $(V_t, E_t)$\\
update ${\bf{D}}_{\tau}$ w.r.t. $({\tilde V}_t, {\tilde E}_t)$ following (\ref{Eq:Neg-tau})\\
generate random noise via $\bm{\Theta} \in \mathbb{R}^{|\tilde V_t| \times d} \sim \mathcal{N} (0, 1/d)$\\
${{\bf{Z}}^{(0)}} \leftarrow {\bm{\Theta}} $ \\
\For{$l$ \bf{from} $1$ \bf{to} $L$}
{
    ${\bf{Z}}^{(l)} \leftarrow {\mathop{\rm ZNorm}\nolimits} (\tanh (\varphi  * {{\bf{Z}}^{(l - 1)}}))$ using (\ref{Eq:Conv}) and (\ref{Eq:Layer})\\
}
${\bf{\tilde Z}} \leftarrow \sigma ({{\bf{Z}}^{(L)}})$ using (\ref{Eq:Layer}) \\
extract ${\bf{\hat Z}}_t \in \mathbb{R}^{|V_t| \times d}$ w.r.t. $V_t$ from ${\bf{\tilde Z}}$ \\
partially fit BIRCH to ${\bf{\hat Z}}_t \in \mathbb{R}^{|V_t| \times d}$\\
${\bf{\hat Z}}_t \in \mathbb{R}^{|{\tilde V}_t| \times d} \leftarrow [{\bf{\hat Z}}_{t-1} ; {\bf{\hat Z}}_t]$ \\
apply BIRCH to ${\bf{\hat Z}}_t$ and get $K$-agnostic $C^{(t)}$ w.r.t. $({\tilde V}_t, {\tilde E}_t)$
\end{algorithm}

According to our analysis about the inference time of InfraredGP (see Table~\ref{Tab:Time}), the downstream BIRCH clustering is the major bottleneck of static GP (i.e., line 7 in Algorithm~\ref{Alg:Static}).
We consider a simple yet effective extension for streaming GP that alleviates such a bottleneck. Concretely, we adopt the online learning mode of BIRCH \cite{zhang1996birch} to partially fit this clustering module to embeddings w.r.t. the newly added topology, instead of running it from scratch in each step.

As stated in Section~\ref{Sec:Prob}, we consider the snowball model for streaming GP.
For each step $t$, let $(V_t, E_t)$ and $({\tilde V_{t-1}}, {\tilde E_{t-1}})$ be the newly added topology and cumulative topology of previous step. We use ${\bf{\hat Z}}_{t-1} \in \mathbb{R}^{|\tilde V_{t-1}| \times d}$ to represent the embeddings derived in previous step w.r.t. $({\tilde V}_{t-1}, {\tilde E}_{t-1})$, which can be re-used for the partial update of BIRCH in current step.

Algorithm~\ref{Alg:Stream} summarizes the extended streaming GP procedure for each step $t > 1$. Given $(V_t, E_t)$ and $({\tilde V_{t-1}}, {\tilde E_{t-1}})$, we first update the cumulative topology $({\tilde V}_t, {\tilde E}_t)$ and degree diagonal matrix ${\bf{D}}_{\tau}$ (i.e., lines 1-2). We then obtain embeddings ${\bf{\tilde Z}} \in \mathbb{R}^{|\tilde V_{t}| \times d}$ w.r.t. $({\tilde V}_t, {\tilde E}_t)$ via one FFP (i.e., lines 3-7), supported by the inductive inference of GNNs \cite{qin2023towards}.
In particular, we extract embeddings ${ \bf{\hat Z} }_t \in \mathbb{R}^{|V_t| \times d}$ w.r.t. the newly added nodes $V_t$ from ${\bf{\tilde Z}} \in \mathbb{R}^{|\tilde V_{t}| \times d}$ (i.e., line 8) and partially update the CFTree of BIRCH only based on ${ \bf{\hat Z} }_t$ (i.e., line 9) via its partial fitting mode \cite{zhang1996birch}.
Finally, we update ${ \bf{\hat Z} }_t$ by appending it to ${ \bf{\hat Z} }_{t-1}$ (i.e., line 10) and apply the updated BIRCH module to ${ \bf{\hat Z} }_t$ to predict a feasible $K$-agnostic partition $C^{(t)}$ w.r.t. the cumulative topology $(\tilde V_t, \tilde E_t)$ (i.e., line 11).

For the initial step $t = 1$, the streaming GP is reduced to the static setting w.r.t. topology $(V_1, E_1)$, so we have to apply Algorithm~\ref{Alg:Static} to $(V_1, E_1)$ and get $(C^{(1)}, {\bf{\hat Z}}_1)$.

Our evaluation of streaming GP (see Fig.~\ref{Fig:Stream-GP-100K} and~\ref{Fig:Stream-GP-1M}) demonstrates that the aforementioned extended procedure (i.e., Algorithm~\ref{Alg:Stream}) can achieve much better efficiency and competitive quality over that of running the static variant (i.e., Algorithm~\ref{Alg:Static}) from scratch in each step.
Surprisingly, the streaming extension may even achieve better quality in some cases.

\section{Experiments}\label{Sec:Exp}
\subsection{Experiment Setups}

\begin{table}[]\scriptsize
\caption{Summary of the Generated Benchmark Datasets}\label{Tab:Data}
\vspace{-0.1cm}
\centering
\begin{tabular}{l|l|l|l|l|l}
\hline
$N$ & $M$ & $K$ & degrees & avg degree & density \\ \hline
5K & 98K-101K & 19 & 4-97 & 39 & 8e-3 \\
10K & 119K-202K & 25 & 3-116 & 40 & 4e-3 \\
50K & 1.01M-1.02M & 44 & 4-106 & 40 & 8e-4 \\
100K & 2.02M-2.03M & 56 & 3-123 & 40 & 4e-4 \\
500K & 10.1M-10.2M & 98 & 2-111 & 40 & 1e-4 \\
1M & 20.3M-20.4M & 125 & 2-126 & 40 & 4e-5 \\ \hline
\end{tabular}
\vspace{-0.4cm}
\end{table}

\begin{table}[]\scriptsize
\caption{Parameter Settings of InfraredGP}\label{Tab:Param}
\vspace{-0.1cm}
\centering
\begin{tabular}{c|l|lll|c|l|lll}
\hline
\multicolumn{1}{l|}{} & $N$ & $\tau$ & $L$ & $d$ & \multicolumn{1}{l|}{} & $N$ & $\tau$ & $L$ & $d$ \\ \hline
\multirow{4}{*}{\textbf{Static}} & 5K & -6 & 10 & 64 & \multirow{2}{*}{\textbf{Static}} & 500K & -80 & 70 & 32 \\
 & 10K & -3 & 9 & 64 &  & 1M & -80 & 60 & 32 \\ \cline{6-10} 
 & 50K & -80 & 40 & 32 & \multirow{2}{*}{\textbf{Streaming}} & 100K & -100 & 20 & 32 \\
 & 100K & -80 & 60 & 32 &  & 1M & -100 & 20 & 16 \\ \hline
\end{tabular}
\vspace{-0.2cm}
\end{table}

\begin{table}[]\scriptsize
\caption{Evaluation Results of \textbf{Static GP} with $N$=5K}\label{Tab:Static-GP-5K}
\vspace{-0.1cm}
\centering
\begin{tabular}{l|l|ll|l}
\hline
\textbf{Methods} & \textbf{Time}($\downarrow$,sec) & \textbf{F1}($\uparrow$,\%) & (PCN,~RCL) & \textbf{ARI}($\uparrow$,\%) \\ \hline
MC-SBM & 98.29$\pm$8.38 & \underline{99.97}$\pm$0.04 & (99.95,~100.00) & \underline{99.97}$\pm$0.04 \\
Par-SBM & \underline{1.76}$\pm$0.14 & \textbf{99.99}$\pm$0.01 & (99.99,~99.99) & \textbf{99.99}$\pm$0.01 \\
Louvain & \underline{1.98}$\pm$0.13 & 97.70$\pm$1.17 & (95.62,~99.90) & 97.53$\pm$1.26 \\
Locale & 2.15$\pm$0.23 & 97.71$\pm$1.10 & (95.57,~99.98) & 97.55$\pm$1.19 \\
RaftGP-C & 5.41$\pm$0.40 & 98.43$\pm$1.35 & (97.08,~99.85) & 98.31$\pm$1.45 \\
RaftGP-M & 5.20$\pm$0.37 & 97.72$\pm$1.82 & (95.71,~99.89) & 97.55$\pm$1.96 \\ \hline
\textbf{InfraredGP} & \textbf{0.11}$\pm$0.03 & \underline{99.92}$\pm$0.09 & (99.99,~99.86) & \underline{99.92}$\pm$0.10 \\
~~Improve. & \textbf{16x} & -0.07\% &  & -0.07\% \\ \hline
\end{tabular}
\vspace{-0.2cm}
\end{table}

\begin{table}[]\scriptsize
\caption{Evaluation Results of \textbf{Static GP} with $N$=10K}\label{Tab:Static-GP-10K}
\vspace{-0.1cm}
\centering
\begin{tabular}{l|l|ll|l}
\hline
\textbf{Methods} & \textbf{Time}($\downarrow$,sec) & \textbf{F1}($\uparrow$,\%) & (PCN,~RCL) & \textbf{ARI}($\uparrow$,\%) \\ \hline
MC-SBM & 259.28$\pm$45.10 & \underline{98.75}$\pm$2.13 & (99.93,~97.68) & \underline{98.67}$\pm$2.25 \\
Par-SBM & \underline{3.48}$\pm$0.35 & \underline{99.71}$\pm$0.49 & (99.47,~99.96) & \underline{99.69}$\pm$0.52 \\
Louvain & 4.88$\pm$0.41 & 95.02$\pm$1.66 & (90.76,~99.74) & 94.70$\pm$1.78 \\
Locale & \underline{4.79}$\pm$0.29 & 96.09$\pm$1.66 & (92.55,~99.96) & 95.84$\pm$1.79 \\
RaftGP-C & 16.70$\pm$0.80 & 98.16$\pm$1.90 & (96.57,~99.88) & 98.04$\pm$2.03 \\
RaftGP-M & 16.32$\pm$0.47 & 98.70$\pm$1.55 & (97.61,~99.85) & 98.61$\pm$1.66 \\ \hline
\textbf{InfraredGP} & \textbf{0.19}$\pm$0.04 & \textbf{99.75}$\pm$0.16 & (99.97, 99.53) & \textbf{99.73}$\pm$0.17 \\
~~Improve. & \textbf{18x} & \textbf{+0.04\%} &  & \textbf{+0.04\%} \\ \hline
\end{tabular}
\vspace{-0.2cm}
\end{table}

\begin{table}[]\scriptsize
\caption{Evaluation Results of \textbf{Static GP} with $N$=50K}\label{Tab:Static-GP-50K}
\vspace{-0.1cm}
\centering
\begin{tabular}{l|l|ll|l}
\hline
\textbf{Methods} & \textbf{Time}($\downarrow$,sec) & \textbf{F1}($\uparrow$,\%) & (PCN,~RCL) & \textbf{ARI}($\uparrow$,\%) \\ \hline
MC-SBM & 2256.41$\pm$247.11 & \underline{99.29}$\pm$0.84 & (99.97,~98.64) & \underline{99.27}$\pm$0.87 \\
Par-SBM & \underline{26.91}$\pm$3.42 & \textbf{99.76}$\pm$0.29 & (99.53,~99.99) & \textbf{99.75}$\pm$0.30 \\
Louvain & \underline{32.68}$\pm$5.74 & 71.85$\pm$5.76 & (56.39,~99.93) & 70.73$\pm$6.09 \\
Locale & 36.23$\pm$1.15 & 93.39$\pm$1.74 & (87.68,~99.96) & 93.17$\pm$1.81 \\
RaftGP-C & 68.82$\pm$2.11 & 98.94$\pm$0.81 & (98.06,~99.84) & 98.90$\pm$0.83 \\
RaftGP-M & 69.43$\pm$3.56 & 99.24$\pm$0.74 & (98.60,~99.90) & 99.21$\pm$0.76 \\ \hline
\textbf{InfraredGP} & \textbf{1.25}$\pm$0.04 & \underline{99.34}$\pm$0.15 & (99.70,~98.97) & \underline{99.32}~0.15 \\
~~Improve. & \textbf{22x} & -0.42\% &  & -0.43\% \\ \hline
\end{tabular}
\vspace{-0.2cm}
\end{table}

\begin{table}[]\scriptsize
\caption{Evaluation Results of \textbf{Static GP} with $N$=100K}\label{Tab:Static-GP-100K}
\vspace{-0.1cm}
\centering
\begin{tabular}{l|l|ll|l}
\hline
\textbf{Methods} & \textbf{Time}($\downarrow$,sec) & \textbf{F1}($\uparrow$,\%) & (PCN,~RCL) & \textbf{ARI}($\uparrow$,\%) \\ \hline
MC-SBM & 5677.10$\pm$882.23 & \underline{99.53}$\pm$0.53 & (99.93,~99.13) & \underline{99.51}$\pm$0.55 \\
Par-SBM & \underline{61.83}$\pm$5.36 & \textbf{99.69}$\pm$0.30 & (99.39,~99.99) & \textbf{99.68}$\pm$0.31 \\
Louvain & 89.47$\pm$10.89 & 53.49$\pm$9.41 & (37.10,~99.94) & 51.78$\pm$9.88 \\
Locale & \underline{87.49}$\pm$1.48 & 86.28$\pm$5.56 & (76.30,~99.96) & 85.89$\pm$5.76 \\
RaftGP-C & 181.90$\pm$1.53 & 99.09$\pm$0.54 & (98.27,~99.93) & 99.07$\pm$0.56 \\
RaftGP-M & 185.73$\pm$6.58 & 99.25$\pm$0.45 & (98.59,~99.93) & 99.24$\pm$0.46 \\ \hline
\textbf{InfraredGP} & \textbf{2.67}$\pm$0.08 & \underline{99.41}$\pm$0.12 & (99.78,~99.04) & \underline{99.39}$\pm$0.12 \\
~~Improve. & \textbf{23x} & -0.28\% &  & -0.28\% \\ \hline
\end{tabular}
\vspace{-0.2cm}
\end{table}

\begin{table}[]\scriptsize
\caption{Evaluation Results of \textbf{Static GP} with $N$=500K}\label{Tab:Static-GP-500K}
\vspace{-0.1cm}
\centering
\begin{tabular}{l|l|ll|l}
\hline
\textbf{Methods} & \textbf{Time}($\downarrow$,sec) & \textbf{F1}($\uparrow$,\%) & (PCN,~RCL) & \textbf{ARI}($\uparrow$,\%) \\ \hline
MC-SBM & OOT & OOT & OOT & OOT \\
Par-SBM & \underline{361.03}$\pm$20.99 & \underline{98.95}$\pm$0.71 & (97.93,~99.99) & \underline{98.93}$\pm$0.71 \\
Louvain & \underline{539.76}$\pm$94.58 & 20.71$\pm$3.19 & (11.59,~99.92) & 18.76$\pm$3.30 \\
Locale & 581.69$\pm$13.64 & \underline{42.49}$\pm$4.60 & (27.09,~99.97) & \underline{41.25}$\pm$4.76 \\
RaftGP-C & OOM & OOM & OOM & OOM \\
RaftGP-M & OOM & OOM & OOM & OOM \\ \hline
\textbf{InfraredGP} & \textbf{16.29}$\pm$1.15 & \textbf{99.42}$\pm$0.09 & (99.94,~98.91) & \textbf{99.41}$\pm$0.09 \\
~~Improve. & \textbf{22x} & \textbf{+0.48\%} &  & \textbf{+0.45\%} \\ \hline
\end{tabular}
\vspace{-0.2cm}
\end{table}

\begin{table}[]\scriptsize
\caption{Evaluation Results of \textbf{Static GP} with $N$=1M}\label{Tab:Static-GP-1M}
\vspace{-0.1cm}
\centering
\begin{tabular}{l|l|ll|l}
\hline
\textbf{Methods} & \textbf{Time}($\downarrow$,sec) & \textbf{F1}($\uparrow$,\%) & (PCN,~RCL) & \textbf{ARI}($\uparrow$,\%) \\ \hline
MC-SBM & OOT & OOT & OOT & OOT \\
Par-SBM & \underline{942.13}$\pm$38.11 & \textbf{99.48}$\pm$0.33 & (98.98,~99.99) & \textbf{99.48}$\pm$0.33 \\
Louvain & \underline{1364.08}$\pm$124.44 & 15.42$\pm$2.09 & (8.37,~99.93) & 13.78$\pm$2.18 \\
Locale & 1457.58$\pm$58.44 & \underline{25.27}$\pm$2.65 & (14.49,~99.96) & \underline{23.90}$\pm$2.72 \\
RaftGP-C & OOM & OOM & OOM & OOM \\
RaftGP-M & OOM & OOM & OOM & OOM \\ \hline
\textbf{InfraredGP} & \textbf{36.67}$\pm$1.11 & \underline{99.40}$\pm$0.10 & (99.96,~98.85) & \underline{99.39}$\pm$0.10 \\
~~Improve. & \textbf{26x} & -0.08\% &  & -0.09\% \\ \hline
\end{tabular}
\vspace{-0.2cm}
\end{table}

\begin{table}[]\scriptsize
\caption{Detailed inference time (s) of \textbf{InfraredGP} in static GP}\label{Tab:Time}
\vspace{-0.1cm}
\centering
\begin{tabular}{l|lll|l|lll}
\hline
$N$ & \textbf{Total} & \textbf{Emb} & \textbf{Clus} & $N$ & \textbf{Total} & \textbf{Emb} & \textbf{Clus} \\ \hline
5K & 0.1076 & 0.0406 & 0.067 & 100K & 2.6689 & 0.4097 & 2.2593 \\
10K & 0.188 & 0.0469 & 0.1411 & 500K & 16.2894 & 1.8373 & 14.4521 \\
50K & 1.2478 & 0.1365 & 1.1113 & 1M & 36.6711 & 3.6771 & 32.9939 \\ \hline
\end{tabular}
\vspace{-0.2cm}
\end{table}

\textbf{Datasets}.
We adopted the IEEE HPEC Graph Challenge benchmark \cite{kao2017streaming} to verify the effectiveness of \textbf{InfraredGP} for both static and streaming GP. In particular, we used the hardest setting of this benchmark, with the (\romannumeral1) block size heterogeneity and (\romannumeral2) ratio between the numbers of within- and between-block edges set to $3$ and $2.5$. The official generator of the benchmark using graph-tool 2.97 was used to generate test graphs, where we set the number of nodes $N$ to be $5$K, $10$K, $50$K, $100$K, $500$K, and $1$M, respectively. For each setting of $N$, we generated five independent graphs and reported the average evaluation results. Statistics of the generated graphs are summarized in Table~\ref{Tab:Data}, where $M$ and $K$ denote the numbers of edges and clusters.

\textbf{Baselines}.
We compared \textbf{InfraredGP} over six baselines, including \textit{MC-SBM} \cite{peixoto2014efficient}, \textit{Par-SBM} \cite{peng2015scalable}, \textit{Louvain} \cite{blondel2008fast}, \textit{Locale} \cite{wang2020community}, \textit{RaftGP-C} \cite{gao2023raftgp}, and \textit{RaftGP-M} \cite{gao2023raftgp}.

Concretely, \textit{MC-SBM} is the standard baseline in Graph Challenge.
\textit{RaftGP-C} and \textit{RaftGP-M} are two variants of \textit{RaftGP}, the innovation
award winner of Graph Challenge 2023.
Similar to \textbf{InfraredGP}, \textit{RaftGP} adopts GCN \cite{kipf2016semi}, a spectral GNN using low-pass filters but with $\tau = 0$, as its backbone, and also feeds random noise to this backbone.

\textbf{Evaluation Protocols}.
Following the evaluation criteria of Graph Challenge, we used \textit{precision}, \textit{recall}, and \textit{adjusted rand index} (ARI) as quality metrics of GP. Based on precision and recall, we also reported the corresponding \textit{F1-score}. The overall \textit{inference time} (sec) of a method to derive a GP result was adopted as the efficiency metric.
We defined that a method encounters the out-of-time (OOT)
exception if it fails to derive a feasible result within $10^4$ seconds.
Usually, larger F1-score, precision, recall, and ARI indicate better GP quality. Smaller inference time implies better efficiency.

\textbf{Parameter \& Environment Settings}.
Table~\ref{Tab:Param} summarizes the parameter settings of \textbf{InfraredGP} on each dataset, where $\tau$, $L$, and $d$ are the (negative) degree correction, number of GNN layers (i.e., iterations), and embedding dimensionality.

All the experiments were conducted on a server with one Intel Xeon Gold 6430 CPU, 120 GB main memory, one RTX4090 GPU (24GB memory), and Ubuntu 22.04 Linux OS.
We used Python 3.10 to develop \textbf{InfraredGP}, where we implemented its GNN backbone using PyTorch 2.1.2 and directly invoked BIRCH functions of scikit-learn 1.6.1 for the downstream clustering.
We adopted the official or widely-used (C/C++ or Python) implementations of all the baselines and tuned their parameters to report the best quality.

\subsection{Evaluation of Static GP}

Evaluation results of static GP for $N$ = $5$K, $10$K, $50$K, $100$K, $500$K, and $1$M are depicted in Tables~\ref{Tab:Static-GP-5K}, \ref{Tab:Static-GP-10K}, \ref{Tab:Static-GP-50K}, \ref{Tab:Static-GP-100K}, \ref{Tab:Static-GP-500K}, and \ref{Tab:Static-GP-1M}, where a metric is in \textbf{bold} or \underline{underlined} if it performs the best or within top-$3$; `OOM' denotes the out-of-memory exception.
Table~\ref{Tab:Time} gives the detailed inference time for different phases of \textbf{InfraredGP}, with `Emb' and `Clus' as the time of embedding derivation and BIRCH clustering.

On all the datasets, \textbf{InfraredGP} can ensure significant improvement of efficiency (i.e., from 16x to 23x) over the second-best baseline. Note that our method derives embeddings via just one FFP (i.e., without any training) using random noise inputs. Surprisingly, its quality degradation w.r.t. the best competitor can always be controlled within $0.5$\%. In some cases, \textbf{InfraredGP} can even achieve slightly better quality. Therefore, \textbf{InfraredGP} ensures a significantly better trade-off between the quality and efficiency of static GP.

Although \textit{RaftGP} adopts an architecture similar to \textbf{InfraredGP}, it still relies on a task-specific hierarchical model selection module and may easily suffer from low efficiency and the OOM exception.
It implies that the simple application of traditional low-pass filters with $\tau = 0$ may fail to derive distinguishable embeddings for some standard clustering algorithms (e.g., BIRCH).
In contrast, \textbf{InfraredGP} can address this limitation by introducing a negative correction mechanism with $\tau < 0$ and amplifying the infrared information.

According to Table~\ref{Tab:Time}, the downstream BIRCH clustering is the major bottleneck of \textbf{InfraredGP}. We further introduce a simple yet effective extension for streaming GP (i.e., Algorithm~\ref{Alg:Stream}) that alleviates this bottleneck.

\begin{figure}[]
\centering
 \begin{minipage}{0.325\linewidth}
 \subfigure[Time$\downarrow$ (sec)]{
  \includegraphics[width=\textwidth,trim=0 0 15 25,clip]{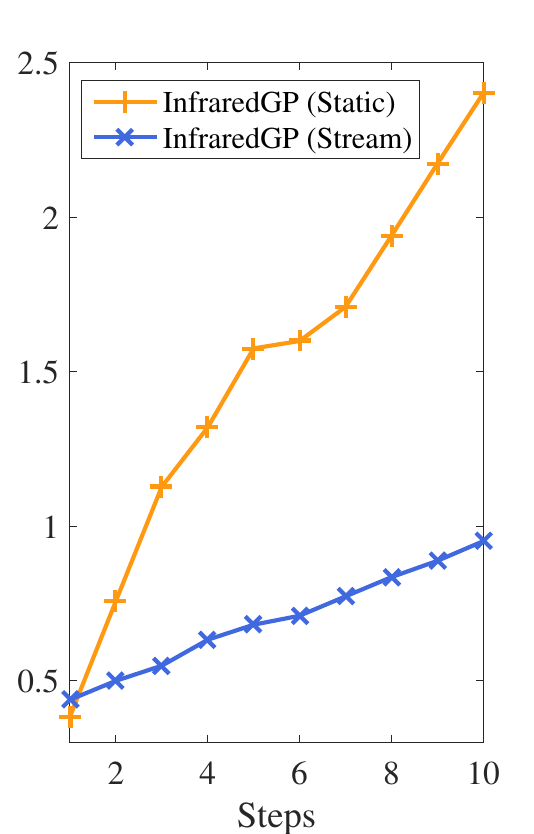}}
 \end{minipage}
 \begin{minipage}{0.325\linewidth}
 \subfigure[F1$\uparrow$]{
  \includegraphics[width=\textwidth,trim=0 0 15 25,clip]{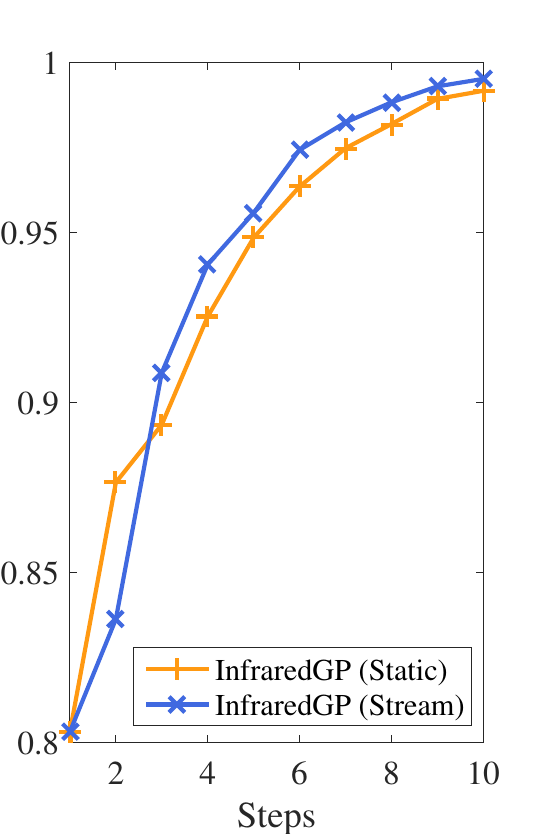}}
 \end{minipage}
 \begin{minipage}{0.325\linewidth}
 \subfigure[ARI$\uparrow$]{
  \includegraphics[width=\textwidth,trim=0 0 15 25,clip]{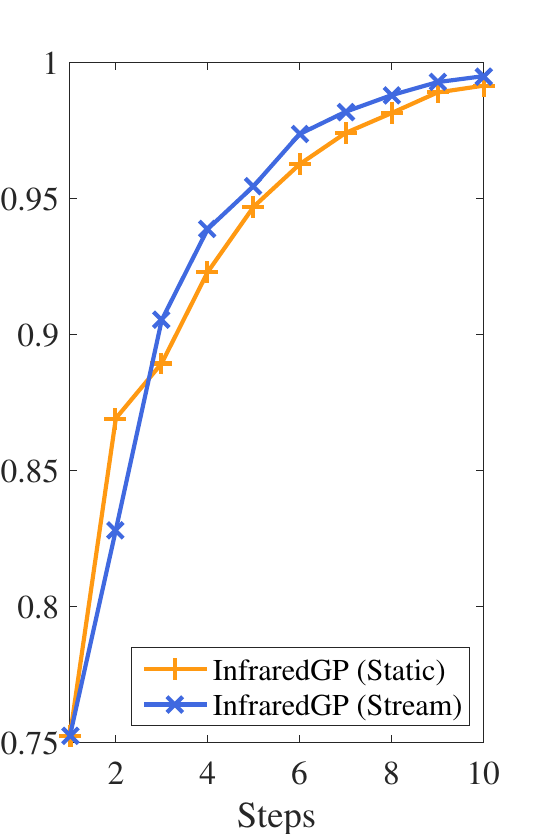}}
 \end{minipage}
\caption{Evaluation results of \textbf{streaming GP} with $N$ = 100K.}\label{Fig:Stream-GP-100K}
\vspace{-0.2cm}
\end{figure}

\begin{figure}[]
\centering
 \begin{minipage}{0.325\linewidth}
 \subfigure[Time$\downarrow$ (sec)]{
  \includegraphics[width=\textwidth,trim=0 0 15 25,clip]{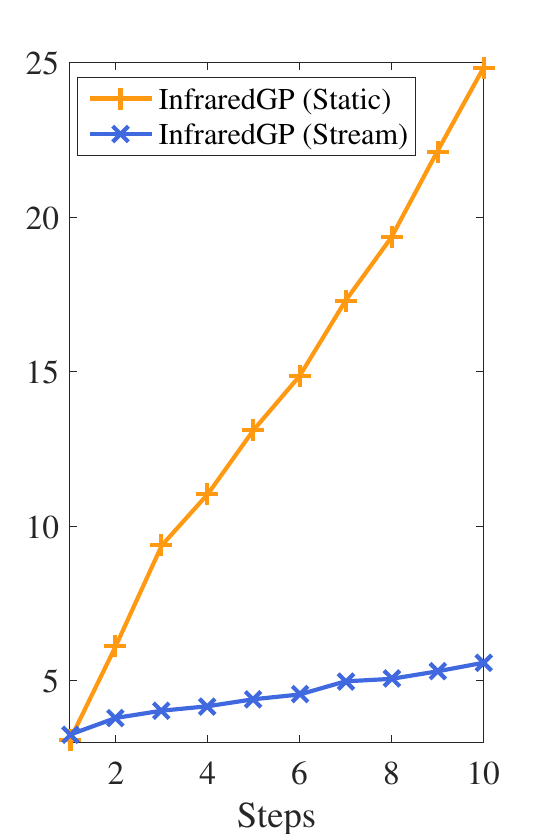}}
 \end{minipage}
 \begin{minipage}{0.325\linewidth}
 \subfigure[F1$\uparrow$]{
  \includegraphics[width=\textwidth,trim=0 0 15 25,clip]{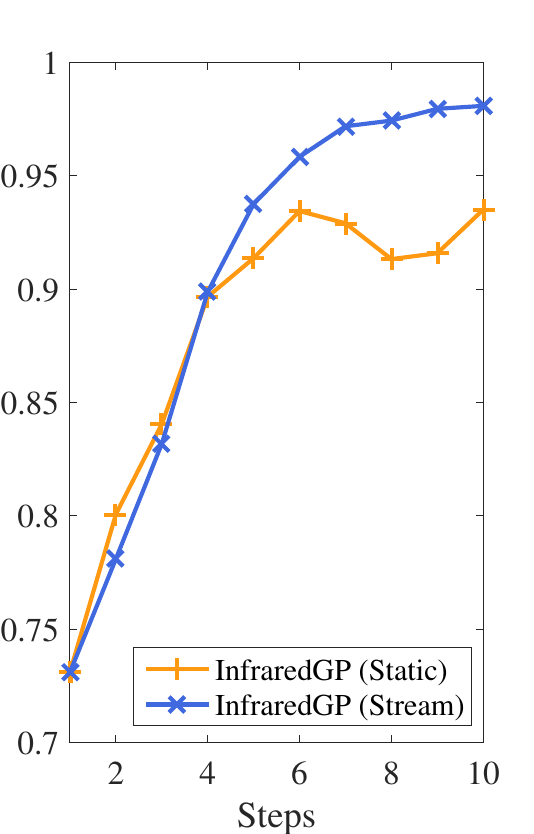}}
 \end{minipage}
 \begin{minipage}{0.325\linewidth}
 \subfigure[ARI$\uparrow$]{
  \includegraphics[width=\textwidth,trim=0 0 15 25,clip]{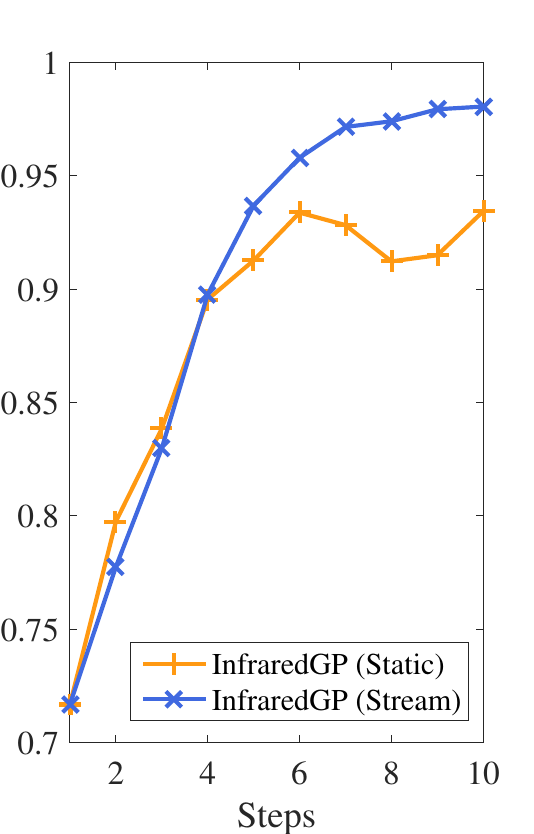}}
 \end{minipage}
\caption{Evaluation results of \textbf{streaming GP} with $N$ = 1M.}\label{Fig:Stream-GP-1M}
\vspace{-0.4cm}
\end{figure}

\subsection{Evaluation of Streaming GP}

We evaluate the extension for streaming GP (i.e., Algorithm~\ref{Alg:Stream}) by comparing its efficiency and quality with that of running the static GP procedure (i.e., Algorithm~\ref{Alg:Static}) from scratch.
As stated in Section~\ref{Sec:Prob}, the snowball model was used to simulate streaming GP. We set the number of steps $T = 10$. The average evaluation results over five independently generated graphs with $N = 100$K and $1$M are shown in Fig.~\ref{Fig:Stream-GP-100K} and \ref{Fig:Stream-GP-1M}, where `static' and `stream' denote running the static and streaming variants (i.e., Algorithms~\ref{Alg:Static} and~\ref{Alg:Stream}).

With the increase of step $t$, the runtime of both variants grows linearly, where the streaming extension can always ensure much better efficiency over the static version for $t \ge 2$. Surprisingly, the streaming extension can even achieve better quality than the static version in some cases. In summary, \textbf{InfraredGP} can be easily extended to support streaming GP.

\begin{figure}[]
  \centering
  \includegraphics[width=\linewidth,trim=51 0 45 5,clip]{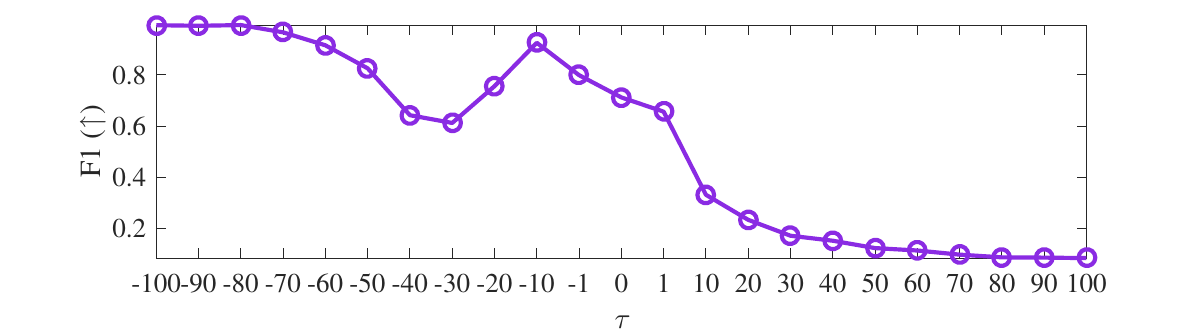}
 \vspace{-0.4cm}
  \caption{Parameter analysis of $\tau$ with $N$=$100$K in terms of F1-score.}\label{Fig:Param-tau}
  \vspace{-0.4cm}
\end{figure}

\subsection{Parameter Analysis \& Ablation Study}
We also verified the degree correction mechanism in \textbf{InfraredGP} by setting $\tau \in \{ -100, -90, \cdots, -10, -1 \}$ (i.e., negative corrections), $\tau = 0$ (i.e., without correction), and $\tau \in \{ 1, 10, \cdots, 100 \}$ (i.e., positive corrections). Example results on the dataset with $N$=$100$K for static GP in terms of F1-score are visualized in Fig.~\ref{Fig:Param-tau}.
Only when $\tau < 0$, \textbf{InfraredGP} can achieve high-quality metrics close to those in Table~\ref{Tab:Static-GP-100K}.
Whereas, all the cases with $\tau \ge 0$ suffer from poor quality.
These results further validate the potential of infrared information (derived by $\tau < 0$) to encode more informative community-preserving embeddings for GP.

\section{Conclusion}\label{Sec:Conc}
In this paper, we followed the IEEE HPEC Graph Challenge to consider $K$-agnostic GP. Based on our investigation about the negative correction mechanism with $\tau < 0$ and graph frequencies $\{ \tilde \lambda_s \}$ beyond the conventional range $[0, 2]$, we proposed InfraredGP.
It (\romannumeral1) adopts a spectral GNN as its backbone combined with low-pass filters and the negative correction mechanism, (\romannumeral2) only feeds random noise to this backbone, and (\romannumeral3) derives embeddings via just one FFP without training. The derived embeddings are then used as the inputs of BIRCH, an efficient $K$-agnostic clustering algorithm, to obtain feasible GP results for Graph Challenge.
Surprisingly, we found out that only based on the negative correction mechanism that amplifies infrared information, our method can derive distinguishable embeddings for some standard clustering modules (e.g., BIRCH) and ensure high inference quality for $K$-agnostic GP. Experiments also demonstrated that InfraredGP can achieve much better efficiency for both static and streaming GP without significant quality degradation.

In this study, only the snowball model of streaming GP was considered. We plan to consider the extension to another emerging edge model of Graph Challenge in our future work.
To extend InfraredGP to attributed graphs \cite{qin2018adaptive,chunaev2020community,qin2021dual,zhao2022trade} and dynamic graphs \cite{qin2023high,qin2023temporal} is also our next research focus.

\section*{Acknowledgment}
This research was supported by the National Natural Science Foundation of China under Grant 62388101, National Science and Technology Major Project (2024ZD01NL00103), Major Key Project of PCL (PCL2025A03), and Interdisciplinary Frontier Research Project of PCL (2025QYB015).


\bibliographystyle{IEEEtran}

\end{document}